\documentclass[runningheads, twocolumns]{llncs}
\usepackage[utf8]{inputenc}
\usepackage[table,xcdraw]{xcolor}

\usepackage{amsmath}
\usepackage{mathtools}
\usepackage{subcaption}
\usepackage{graphicx}
\usepackage{listings}
\usepackage{gensymb}
\usepackage{algorithm}
\usepackage[noend]{algpseudocode}
\usepackage[shortlabels]{enumitem}
\usepackage{hyperref}
\usepackage{tabto}
\usepackage{multicol}
\usepackage{todonotes}

\begin{document}

\title{A deep active inference model of the rubber-hand illusion}

%
\author{Thomas Rood \and
Marcel van Gerven \and
Pablo Lanillos}
\authorrunning{T. Rood et al.}
%
\institute{
Department of Artificial Intelligence\\
Donders Insitute for Brain, Cognition and Behaviour\\
Montessorilaan 3, 6525 HR Nijmegen, the Netherlands\\
\email{p.lanillos@donders.ru.nl}}


%
\maketitle              
\begin{abstract}
Understanding how perception and action deal with sensorimotor conflicts, such as the rubber-hand illusion (RHI), is essential to understand how the body adapts to uncertain situations. Recent results in humans have shown that the RHI not only produces a change in the perceived arm location, but also causes involuntary forces. Here, we describe a deep active inference agent in a virtual environment, which we subjected to the RHI, that is able to account for these results. We show that our model, which deals with visual high-dimensional inputs, produces similar perceptual and force patterns to those found in humans.

\end{abstract}

\section{Introduction}
The complex mechanisms underlying perception and action that allow seamless interaction with the environment are largely occluded from our consciousness. To interact with the environment in a meaningful way, the brain must integrate noisy sensory information from multiple modalities into a coherent world model, from which to generate and continuously update an appropriate action~\cite{sensorimotorbayes}. Especially, how the brain-body deals with sensorimotor conflicts~\cite{hinz2018drifting,Lanillos2020.07.08.191304}, e.g., conflicting information from different senses, is an essential question for both cognitive science and artificial intelligence. 
Adaptation to unobserved events and changes in the body and the environment during interaction is a key characteristic of body intelligence that machines still fail at. 

The rubber-hand illusion (RHI)~\cite{ogrhi} is a well-known experimental paradigm from  cognitive science that allows the investigation of body perception under conflicting information in a controlled setup.
During the experiment, human participants cannot see their own hand but rather perceive an artificial hand placed in a different location (e.g. 15 cm from their current hand). After a minute of visuo-tactile stimulation~\cite{rhitime}, the perceived location of the real hand drifts towards the location of the artificial arm and suddenly the new hand becomes part of their own. 

We can find some RHI modelling attempts in the literature; see~\cite{kilteni2015over} for an overview until 2015.
In \cite{samadcausal}, a Bayesian causal inference model was proposed to estimate the perceived hand position after stimulation. In~\cite{hinz2018drifting} a model inspired by the free-energy principle~\cite{unifiedbrain} was used to synthetically test the RHI in a robot. The perceptual drift (mislocalization of the hand) was compared to that of humans observations. 

Recent experiments have shown that humans also  generate meaningful force patterns towards the artificial hand during the RHI~\cite{asai2015illusory,Lanillos2020.07.08.191304}, adding the action dimension to this paradigm. We hypothesise that the strong interdependence between perception and action can be accounted for by mechanisms underlying active inference~\cite{actionandbehaviorfep}.


In this work, we propose a \textit{deep active inference} model of the RHI, based on~\cite{lanillos2018adaptive,oliver2019active,sancaktar2020end}, where an artificial agent directly operates in a 3D virtual reality  (VR) environment\footnote{Code will be publicly available at  \url{https://github.com/thomasroodnl/active-inference-rhi}}. Our model 1) is able to produce similar perceptual and active patterns to human observations during the RHI and 2) provides a scalable approach for further research on body perception and active inference, as it deals with high-dimensional inputs such as visual images originated from the 3D environment.

\section{Deep active inference model} \label{free-energy-model}
\begin{figure*}[hbtp!]
    \centering
    \begin{subfigure}{0.55\textwidth}
    \centering
        \includegraphics[width=0.9\linewidth]{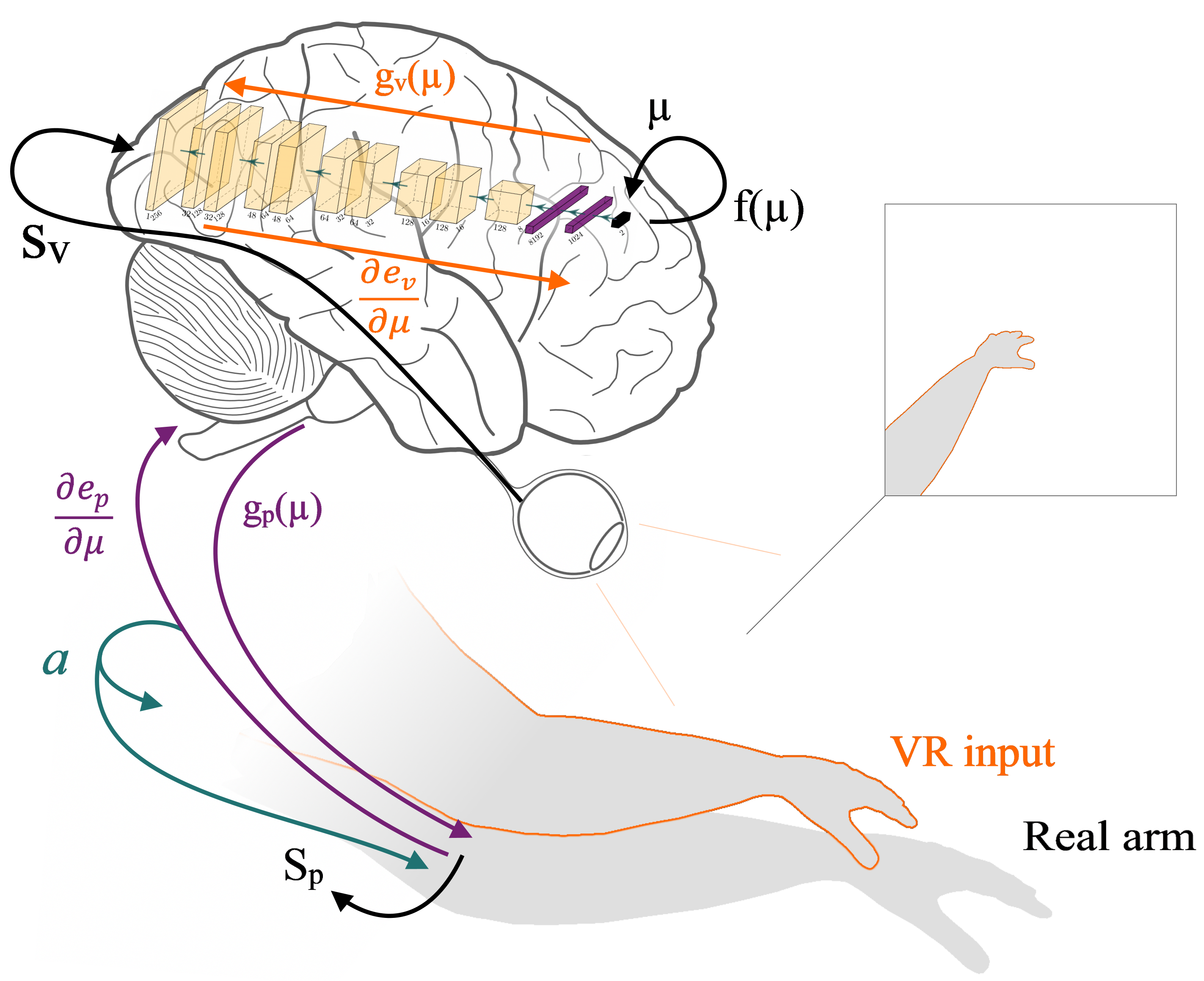}
        \caption{Deep Active inference} \label{fig:modelscheme}
    \end{subfigure}
    \begin{minipage}{0.4\textwidth}
    \begin{subfigure}{\linewidth}
        \includegraphics[width=\linewidth]{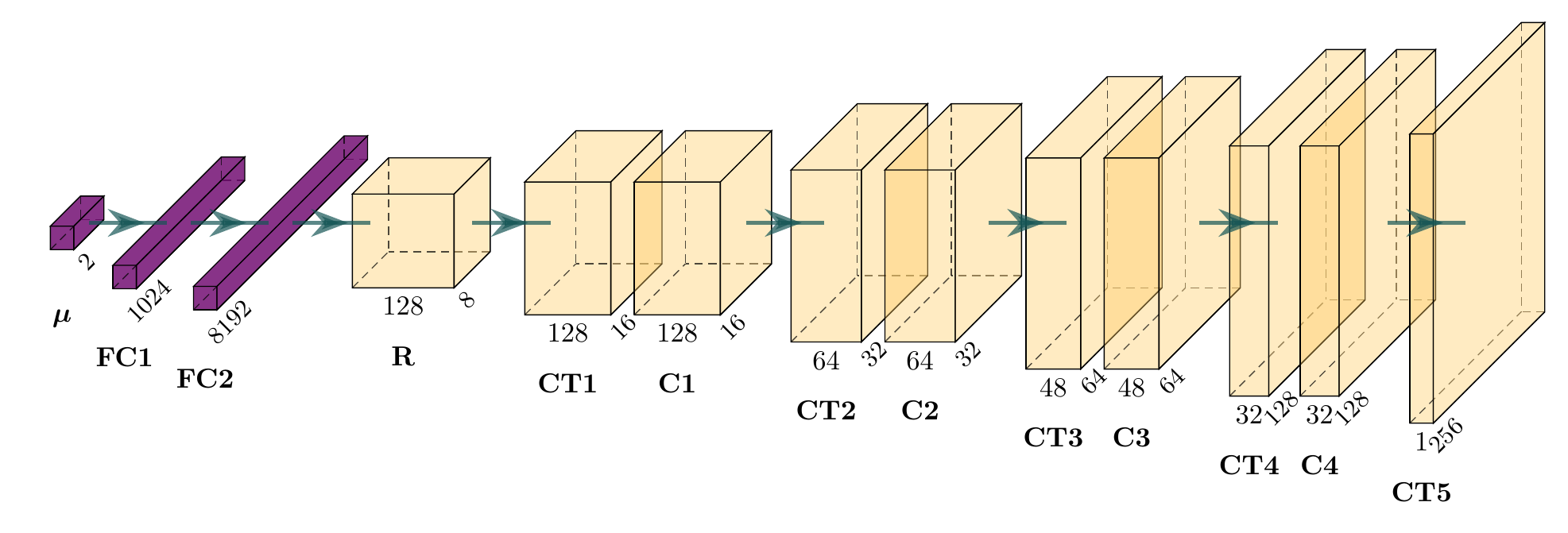}
        \caption{Convolutional decoder} \label{fig:decoder}
    \end{subfigure}
    
    \vspace*{1cm}
    \begin{subfigure}{\linewidth}
        \includegraphics[width=\linewidth]{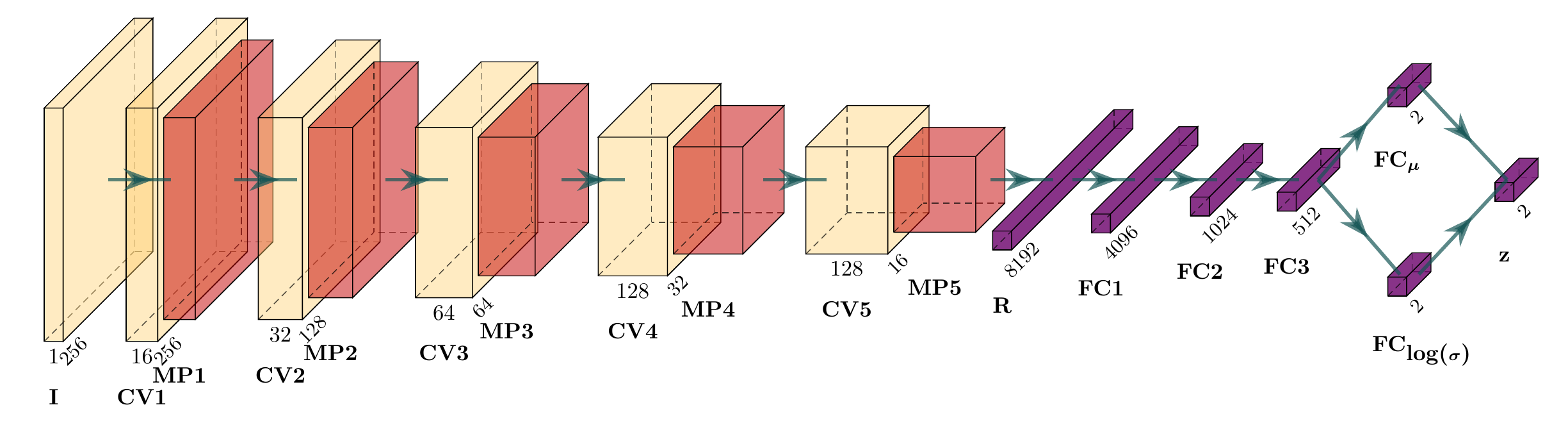}
        \caption{VAE (encoder only)} \label{fig:vae}
    \end{subfigure}
    \end{minipage}
    \caption{Deep active inference model for the virtual rubber-hand illusion. (a)~The brain variables $\mu$ that represent the body state are inferred through proprioceptive $e_p$ and visual $e_v$ prediction errors and their own dynamics $f(\mu)$. During the VR immersion, the agent only sees the VR arm. The ensuing action is driven by proprioceptive prediction errors. The generative visual process is approximated by means of a deep neural network that encodes the sensory input into the body state through a bottleneck. (b,c) Visual generative architectures tested.}
    \label{fig:model}
\end{figure*}

We formalise body perception and action as an inference problem \cite{kappen2012optimal,botvinick2012planning,actionandbehaviorfep,oliver2019active}.
The unobserved body state is inferred from the senses (observations) while taking into account its state prior information. 
To this end, the agent makes use of two sensory modalities. The visual input $s_v$ is described by a pixel matrix (image) and the proprioceptive information $s_p$ represents the angle of every joint of the arm -- See Fig. \ref{fig:modelscheme}.

Computation of the body state is performed by optimizing the the variational free-energy bound~\cite{actionandbehaviorfep,oliver2019active}. Under the mean-field and Laplace approximations and defining $\mu$ as the brain variables that encode the variational density that approximates the body state distribution and defining $a$ as the action exerted by the agent, perception and action are driven by the following system of differential equations (see~\cite{friston2007variational,buckley2017free,sancaktar2020end} for a derivation):
\begin{align}
    \dot{\mu} &= -\partial_\mu  F = - \partial_\mu e_p^T \Sigma_p^{-1} e_p - \partial_\mu e_v^T \Sigma_v^{-1} e_v -\partial_\mu e_f^T \Sigma_\mu^{-1} e_f
    \label{eq:mufep}\\
    \dot{a} &= -\partial_a  F = - \partial_a e_p^T \Sigma_p^{-1} e_p \\
    e_p &= s_p- g_p(\mu) \\
    e_v &= s_v- g_v(\mu) \\
    e_f &= - f(\mu)
    \label{eq:ef}
\end{align}
Note that this model is a specific instance of the full active inference model~\cite{unifiedbrain} tailored to the RHI experiment. We wrote the variational free-energy bound in terms of the prediction error $e$ and for clarity, we split it into three terms that correspond to the visual, proprioceptive and dynamical component of the body state. The variances $\Sigma_v, \Sigma_p, \Sigma_\mu$ encode the reliability of the visual, proprioceptive and dynamics information, respectively, that is used to infer the body state. The dynamics of the prediction errors are governed by different generative processes. Here, $g_v(\mu)$ is the generative process of the visual information (i.e. the predictor of the visual input given the brain state variables), $g_p(\mu)$ is the proprioceptive generative process and $f(\mu)$ denotes internal state dynamics (i.e. how the brain variables evolve in time)\footnote{Note that in Equation \eqref{eq:ef}, the prediction error with respect to the internal dynamics $e_f = \mu' - f(\mu)$ was simplified to $e_f = -f(\mu)$ under the assumption that $\mu' = 0$. In other words, we assume no dynamics on the internal variables.}. 

Due to the static characteristics of the passive RHI experiment we can simplify the model. First, the generative dynamics model does not affect body update because the experimental setup does not allow for body movement. Second, we fully describe the body state by the joint angles. This means that the $s_p$ and the body state match. Thus, $g(\mu) = \mu$ plus noise and the inverse mapping $\partial_\mu g_p(\mu)$ becomes an all-ones vector. Relaxing these two assumptions is out of the scope of this paper. We can finally write the differential equations with the generative models as follows:
\begin{align}
    \dot{\mu}&=   \Sigma_p^{-1}(s_p -g_p(\mu)) 
    + \partial_{\mu} g_v(\mu)^T \gamma \Sigma_v^{-1}(s_v -g_v(\mu)) \label{eq:mufep2} \\
    \dot{a}&= - \Delta_t \Sigma_p^{-1}(s_p -g_p(\mu))
    \label{eq:afep2}
\end{align}
where $\gamma$ has been included in the visual term to modulate the level of causality regarding whether the visual information has been produced by our body in the RHI -- see Sec.~\ref{eq:causal}. Equation \ref{eq:afep2} is only valid if the action is the velocity of the joint. Thus, the sensor change given the action corresponds to the time interval between each iteration $\partial_a s= \Delta_t$.

We scale up the model to high-dimensional inputs such as images by approximating the visual generative model $g_v(\mu)$ and the partial derivative of the error with respect to the brain variables $\partial_\mu e_v$ by means of deep neural networks, inspired by~\cite{sancaktar2020end}.

\subsection{Generative model learning}
We learn the forward and inverse generative process of the sensory input by exploiting the representational capacity of deep neural networks. Although in this work we only address the visual input, this method can be extended to any other modality. To learn the the visual forward model $g_v(\mu)$ we compare two different deep learning architectures, that is, a convolutional decoder (Fig. \ref{fig:decoder}) and a variational autencoder (VAE, Fig. \ref{fig:vae}). 

The convolutional decoder was designed in similar fashion to the architecture used in \cite{sancaktar2020end}. After training the relation between the visual input and the body state, the visual prediction can be computed through the forward pass of the network and its inverse $\partial g(\mu)/\partial \mu$ by means of the backward pass. The VAE was designed using the same decoding structure as the convolutional decoder to allow a fair performance comparison. This means that these models mainly differed in the way they were trained. In the VAE approach we train using the full architecture and we just use the decoder to compute the predictions in the model.

\subsection{Modelling visuo-tactile stimulation synchrony}\label{eq:causal}
To synthetically replicate the RHI we need to model both synchronous and asynchronous visuo-tactile stimulation conditions. We define the timepoints at which a visual stimulation event and the corresponding tactile stimulation take place, denoted $t_v$ and $t_t$ respectively. Inspired by the Bayesian causal model~\cite{samadcausal}, we distinguish between two causal explanations of the observed data. That is, $C = c_1$ signifies that the observed (virtual) hand produced both the visual and the tactile events whereas $C=c_2$ signifies that the observed hand produced the visual event and our real hand produced the tactile event (visual and tactile input come from two different sources). The causal impact of the visual information on the body state is represented by
\small
\begin{equation}\label{eq:8}
\gamma =   p(c_1 \mid t_v, t_t) = \frac{p(t_v, t_t \mid c_1)  p(c_1)}{p(t_v, t_t \mid c_1)  p(c_1) + p(t_v, t_t \mid c_2)  p(c_2)}
\end{equation}
\normalsize
where $p(t_v, t_t \mid c_1)$ is defined as a zero-mean Gaussian distribution over the difference between the timepoints ($p(t_v - t_t \mid c_1)$) and $p(t_v, t_t \mid c_2)$ is defined as a uniform distribution since under $c_2$, no relation between $t_v$ and $t_t$ is assumed. 
This yields the update rule
\begin{equation}\label{eq:11}
  \gamma_{t+1} = 
  \begin{cases}
    \frac{p(t_v, t_t \mid c_1)  \gamma_t}{p(t_v, t_t | \gamma_t) \cdot \gamma_t + p(t_v, t_t \mid c_2) (1-\gamma_t)}  &\text{if visuo-tactile event}\\
    \gamma_t \cdot exp(-\frac{(t - max(t_{v}, t_{t}))^2}{\Delta_t^{-1}} \cdot r_{decay}), &\text{otherwise}
  \end{cases}
\end{equation}
Note that $\gamma$ is updated only in case of visuo-tactile events. Otherwise, an exponential decay is applied.


\begin{figure*}[hbtp!]
    \centering
    \includegraphics[width=0.8\textwidth, height=120px]{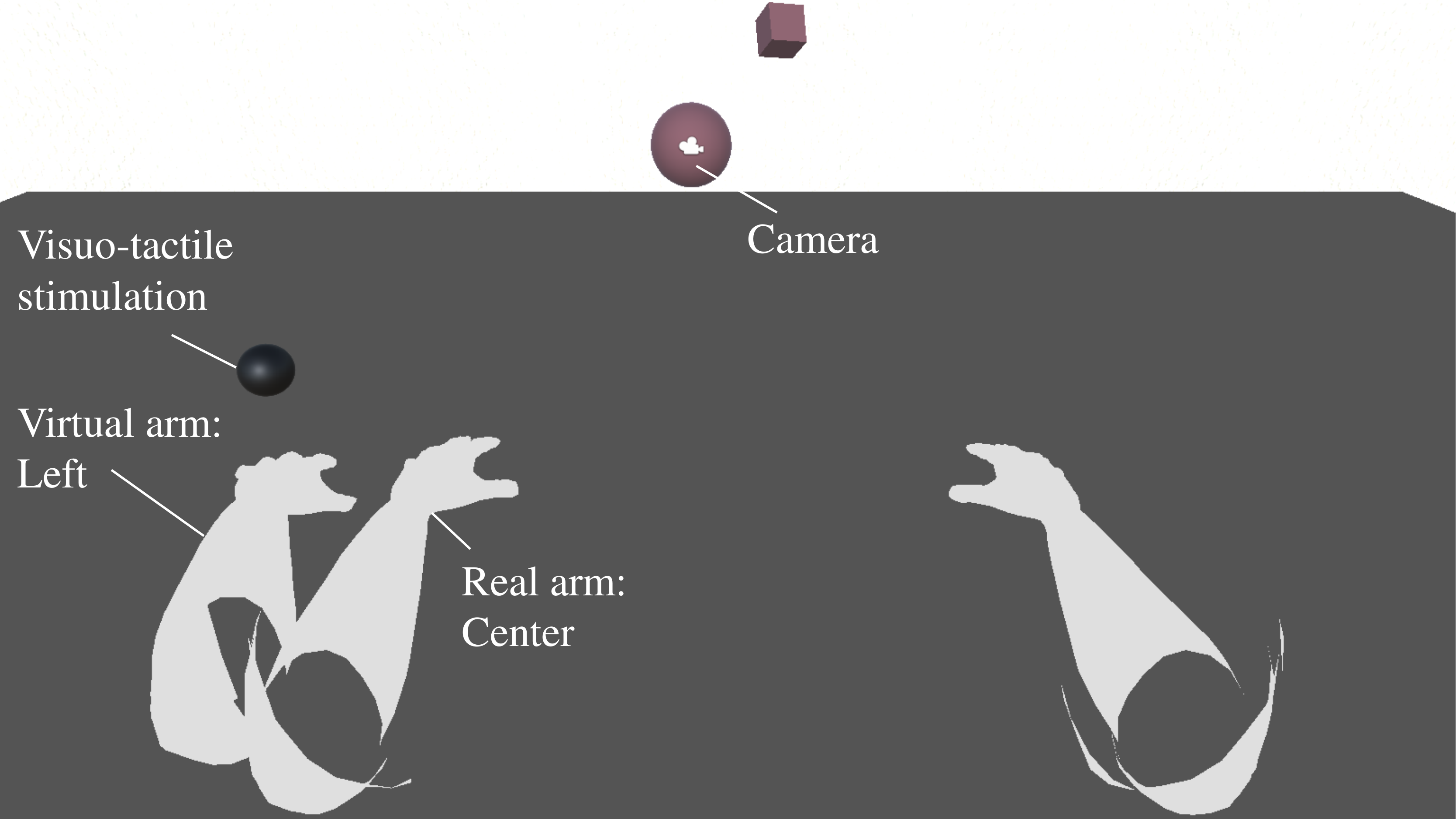}
    \caption{Virtual environment and experimental setup modelled in the Unity engine.}
    \label{fig:setup}
\end{figure*}
\section{Experimental setup}

We modelled the RHI in a virtual environment created in Unity, as depicted in Fig. \ref{fig:setup}. This environment was build to closely match the experimental setup used in the human study described in \cite{Lanillos2020.07.08.191304}. This experiment exposed human participants to a virtual arm located to the left and right of their real arm, and applied visuo-tactile stimulation by showing a virtual ball touching the hand and applying a corresponding vibration to the hand.
Here, the agent's control consisted of two degrees of freedom: shoulder 
adduction/abduction and elbow flexion/extension. The environment provided proprioceptive information on the shoulder and elbow joint angles to the agent. Visual sensory input to the model originated from a camera located between the left and the right eye position, producing $256 \times 256$ pixel grayscale images. Finally, the ML-Agents toolkit was used to interface between the Unity environment and the agent in Python~\cite{juliani2018unity}.
The agent arm was placed in a forward resting position such that the hand was located 30 cm to the left of the body midline (center position). Three virtual arm location conditions were evaluated: Left, Center and Right. The Center condition matched the information given by proprioceptive input. Visuo-tactile stimulation was applied by generating a visual event at a regular interval of two seconds, followed by a tactile event after a random delay sampled in the range [0, 0.1) for synchronous stimulation and in the range [0, 1) for asynchronous stimulation. The initial $\gamma$ value was set to 0.01 and we ran $N = 5$ trials each for 30 s (1500 iterations).
\begin{figure}[b!]
    \centering
    \begin{subfigure}[t]{0.28296\textwidth}
        \includegraphics[width=\linewidth]{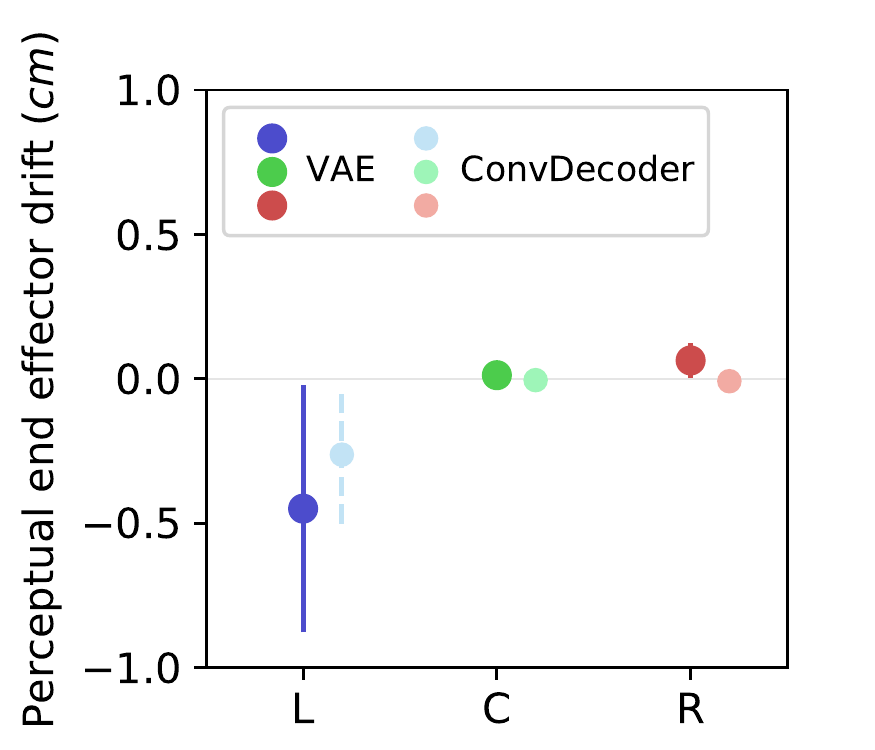}
        \caption{Perception all}\label{fig:rhiperceptualdrifboth}
    \end{subfigure}
    \begin{subfigure}[t]{0.2424\textwidth}
        \includegraphics[width=\linewidth]{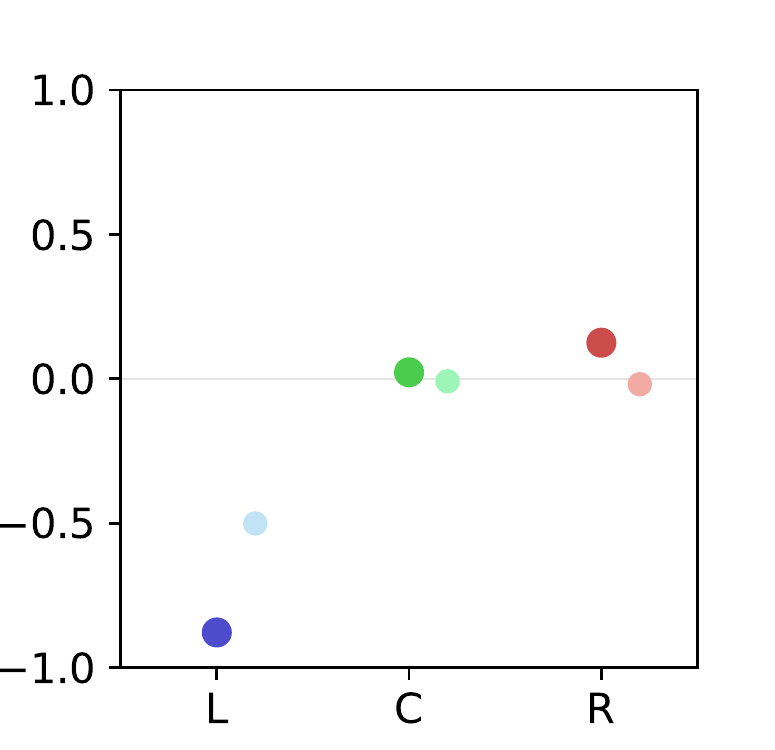}
        \caption{Percept. Sync.}\label{fig:rhiperceptualdrifsync}
    \end{subfigure}
    \begin{subfigure}[t]{0.2424\textwidth}
        \includegraphics[width=\linewidth]{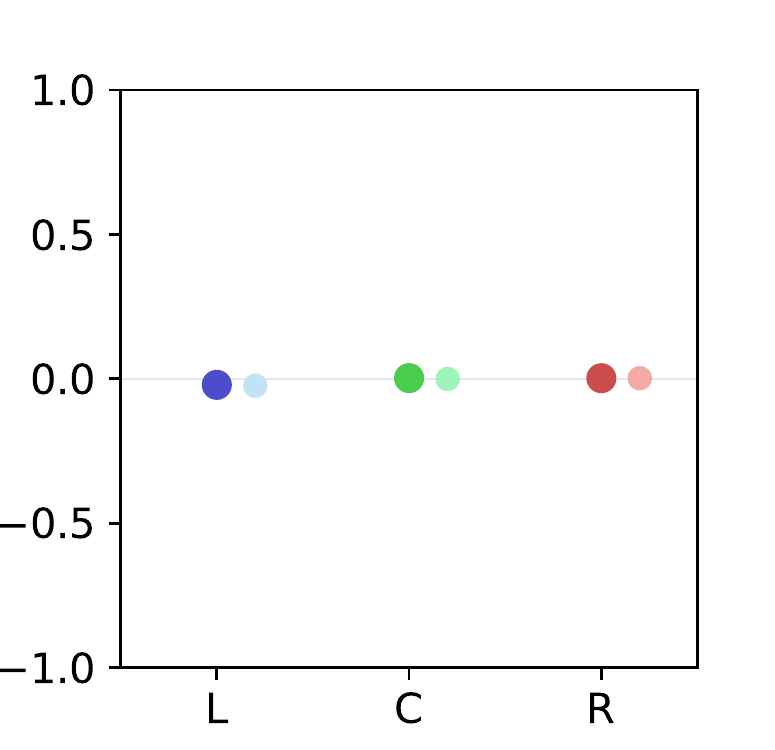}
        \caption{Percept. Async.}\label{fig:rhiperceptualdrifasync}
    \end{subfigure}
    
    \begin{subfigure}[t]{0.28296\textwidth}
        \includegraphics[width=\linewidth]{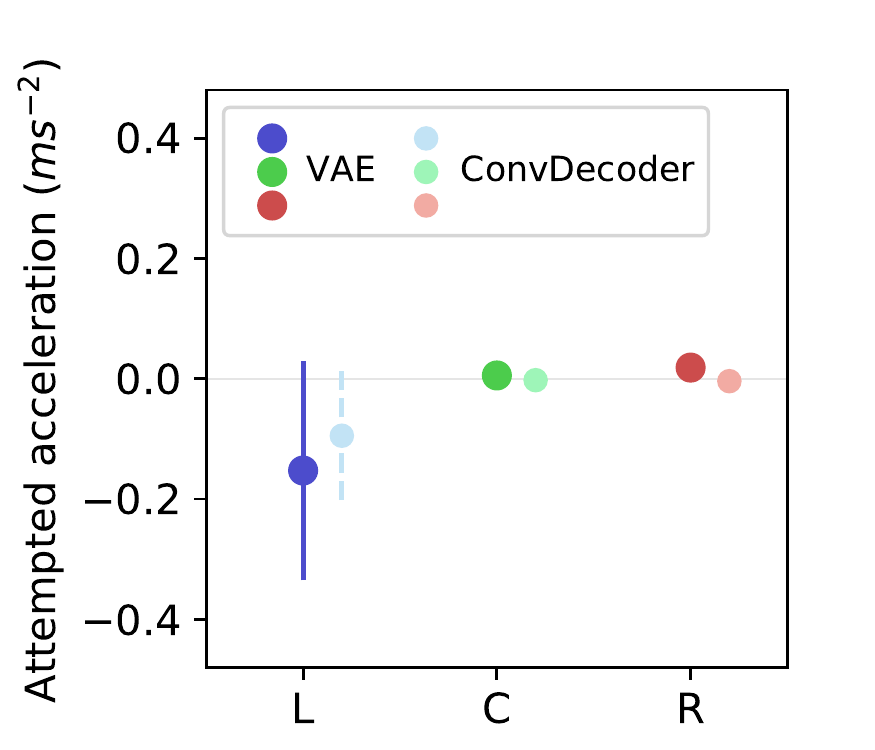}
        \caption{Action all}\label{fig:rhifwdall}
    \end{subfigure}
    \begin{subfigure}[t]{0.2424\textwidth}
        \includegraphics[width=\linewidth]{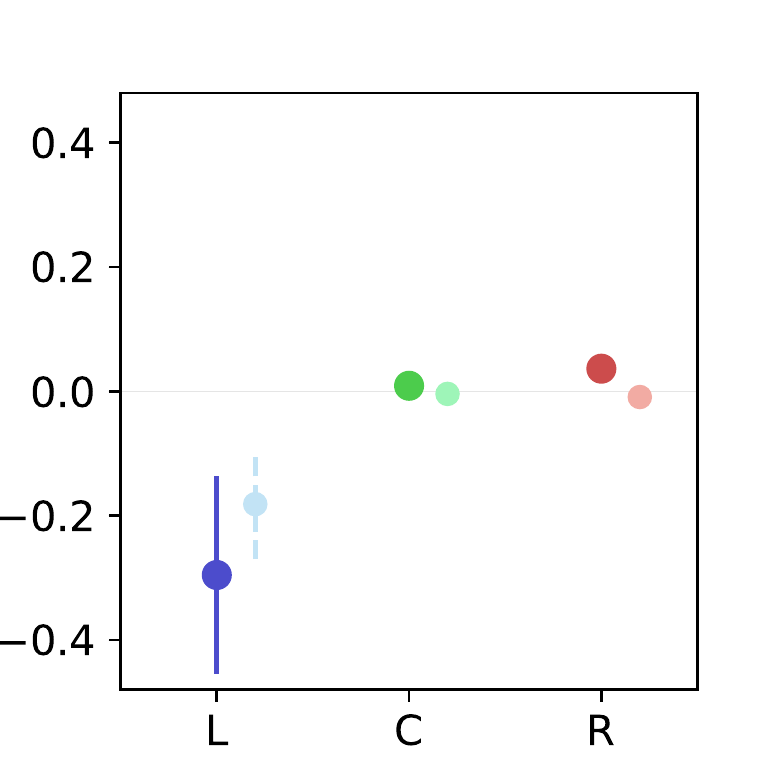}
        \caption{Action Sync.}\label{fig:rhifwdsync}
    \end{subfigure}
    \begin{subfigure}[t]{0.2424\textwidth}
        \includegraphics[width=\linewidth]{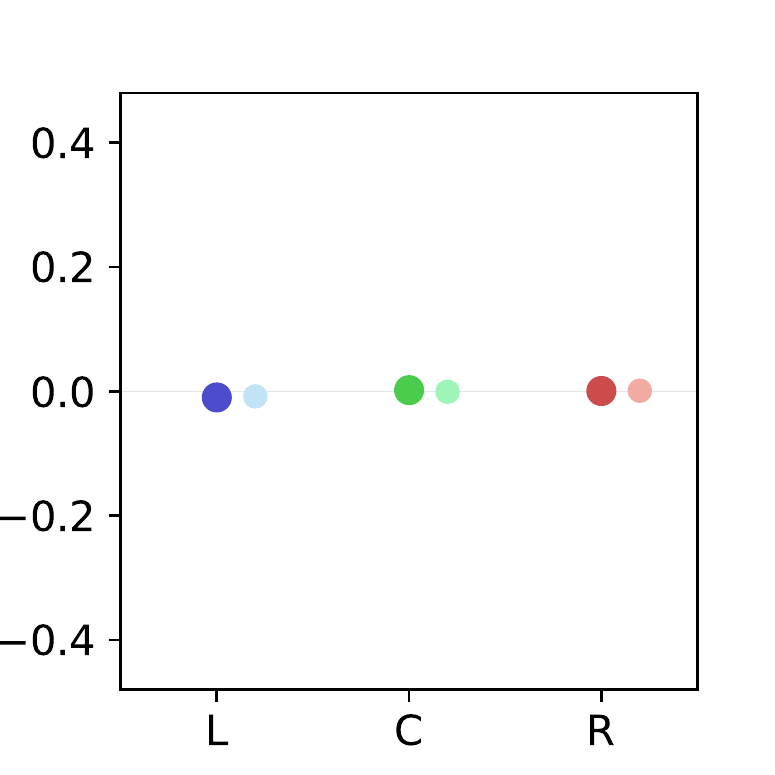}
        \caption{Action Async.}\label{fig:rhifwdasync}
    \end{subfigure}
    \begin{subfigure}[t]{0.80\textwidth}
        \includegraphics[width=\linewidth, height=80px]{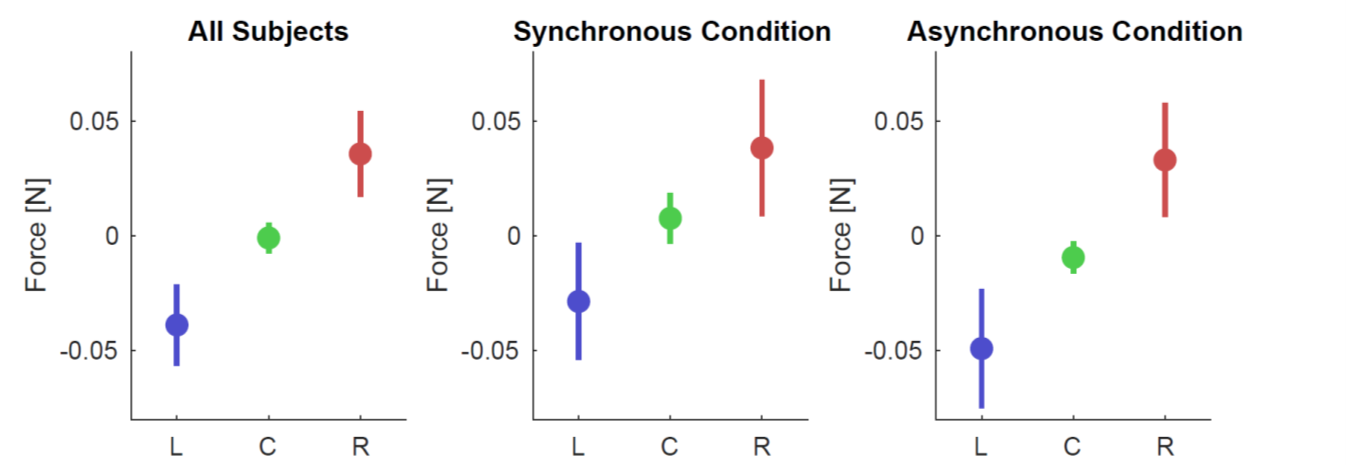}
        \caption{Human recorded forces}\label{fig:rhihuman}
    \end{subfigure}
    
    \begin{subfigure}[t]{0.80\textwidth}
        \includegraphics[width=\linewidth,height=60px]{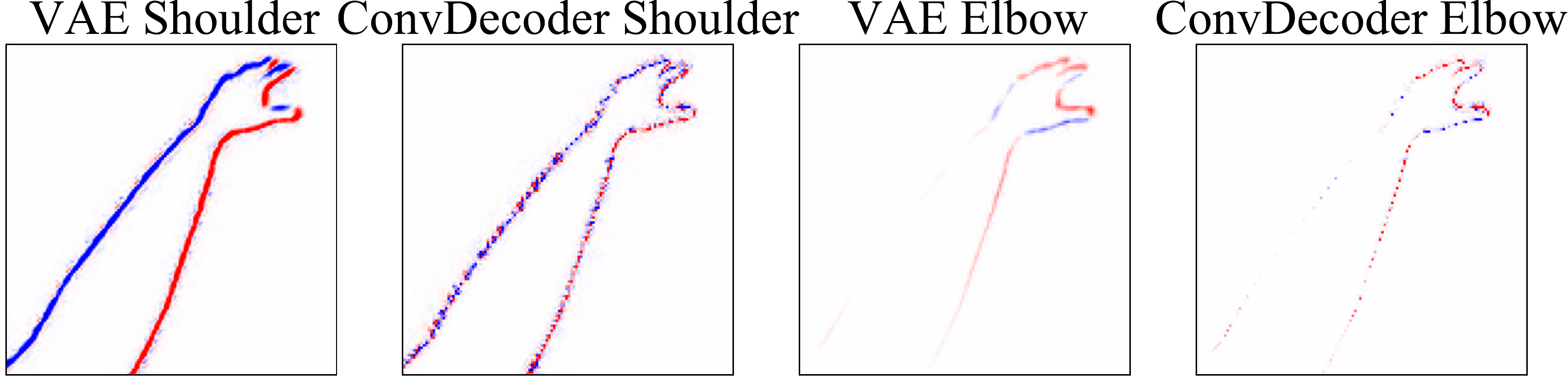}
        \caption{Jacobian $\partial_\mu g(\mu)$ learnt for both visual models.}\label{fig:jacobians}
    \end{subfigure}
    \caption{Model results. (a, b, c) Mean perceptual end-effector drift (in cm). (d,e,f) Mean horizontal end-effector acceleration. (g) Mean forces exerted by human participants in a virtual rubber-hand experiment (from~\cite{Lanillos2020.07.08.191304}). (h) Visual representation of the Jacobian learnt for the visual models.}
    \label{fig:rhifwd}
\end{figure}
\section{Results}
\label{sec:results}
We observed similar patterns in the drift of the perceived end-effector location (Fig. \ref{fig:rhiperceptualdrifboth}) and the end-effector action (Fig. \ref{fig:rhifwd}). These agree with the behavioural data obtained in human experiments (Fig.~\ref{fig:rhihuman}). For the left and right condition, we observed forces in the direction of the virtual hand during synchronous stimulation (Fig. \ref{fig:rhifwdsync}). However, non-meaningful forces were produced using the convolutional decoder for the right condition. For the center condition, both models produced near-zero average forces. Lastly, asynchronous stimulation produced, with both models, attenuated forces (Fig.~\ref{fig:rhifwdasync}). The learnt visual representation differed between the VAE and the Convolutional decoder approaches (Fig. \ref{fig:jacobians}). The VAE obtained smoother and more bounded visual Jacobian values, likely due to its probabilistic latent space.

\section{Conclusion}
In this work, we described a deep active inference model to study body perception and action during sensorimotor conflicts, such as the RHI. The model, operating as an artificial agent in a virtual environment, was able to produce similar perceptual and active patterns to those found in humans. Further research will address how this model can be employed to investigate the construction of the sensorimotor self \cite{lanillos2017enactive}.



\bibliographystyle{splncs04}
\bibliography{references,pl}
\end{document}